\documentclass[10pt,twocolumn,letterpaper]{article}

\usepackage[pagenumbers]{cvpr} 

\usepackage{graphicx}
\usepackage{amsmath}
\usepackage{amssymb}
\usepackage{booktabs}

\usepackage[ruled, lined, linesnumbered, commentsnumbered, longend]{algorithm2e}

\usepackage{comment}
\usepackage{dblfloatfix}
\usepackage{lipsum}

\usepackage[pagebackref,breaklinks,colorlinks]{hyperref}

\usepackage[capitalize]{cleveref}
\crefname{section}{Sec.}{Secs.}
\Crefname{section}{Section}{Sections}
\Crefname{table}{Table}{Tables}
\crefname{table}{Tab.}{Tabs.}

\def\cvprPaperID{11832} 
\def\confName{CVPR}
\def\confYear{2023}

\newcommand{\methodname}{MCMPG}

\newcommand{\graphnameFullsmall}{multi-frame proposal graph}
\newcommand{\graphnameFullcapital}{Multi-frame Proposal Graph}
\newcommand{\graphnameshort}{\textit{MP-Graph}}

\begin{document}

\title{Maximal Cliques on Multi-Frame Proposal Graph for Unsupervised Video Object Segmentation}

\author{Jialin Yuan $\star$ \\
Oregon State University\\
{\tt\small yuanjial@oregonstate.edu}
\and
Jay Patravali $\star$ \\
Oregon State University\\
{\tt\small patravaj@oregonstate.edu}
\and
Hung Nguyen\\
Oregon State University\\
{\tt\small nguyehu5@oregonstate.edu}
\and
Chanho Kim\\
Oregon State University\\
{\tt\small kimchanh@oregonstate.edu}
\and
Li Fuxin\\
 Oregon State University\\
 {\tt\small lif@oregonstate.edu}
}

\maketitle

\newcommand\blfootnote[1]{%
  \begingroup
  \renewcommand\thefootnote{}\footnote{#1}%
  \addtocounter{footnote}{-1}%
  \endgroup
}
\blfootnote{$\star$ Equal contributions.}

\begin{abstract}
   Unsupervised Video Object Segmentation (UVOS) aims at discovering objects and tracking them through videos. For accurate UVOS, we observe if one can locate precise segment proposals on key frames, subsequent processes are much simpler. Hence, we propose to reason about key frame proposals using a graph built with the object probability masks initially generated from multiple frames around the key frame and then propagated to the key frame. On this graph, we compute maximal cliques, with each clique representing one candidate object. By making multiple proposals in the clique to vote for the key frame proposal, we obtain refined key frame proposals that could be better than any of the single-frame proposals. A semi-supervised VOS algorithm subsequently tracks these key frame proposals to the entire video. Our algorithm is modular and hence can be used with any instance segmentation and semi-supervised VOS algorithm. We achieve state-of-the-art performance on the DAVIS-2017 validation and test-dev dataset. On the related problem of video instance segmentation, our method shows competitive performance with the previous best algorithm that requires joint training with the VOS algorithm. 
\end{abstract}

\section{Introduction}

\label{sec:intro}
\begin{figure*}[ht]
  \centering
  \includegraphics[width=1.0\textwidth]{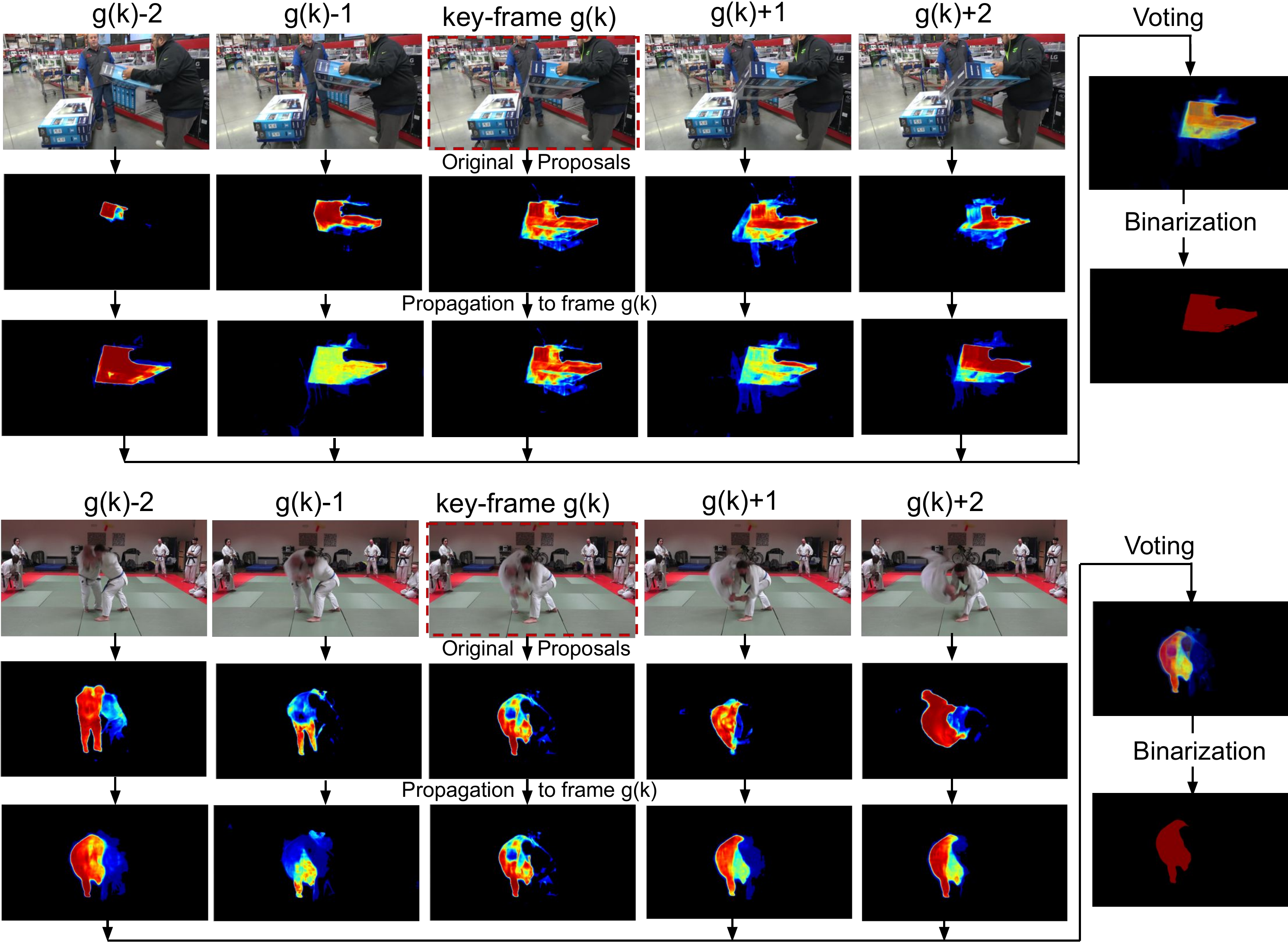}
  \caption{Illustration of the proposed {\methodname} for object proposal refinement using a key frame clip with size $5$ ( \textsl{"loading"} and \textsl{"judo"} from DAVIS 2017 \textsl{val} set). 
 \textbf{On the left side:} the first row are the RGB frames within the local window (the key frame $g(k)$ is the central frame highlighted in the red border). The second row is the segmentation of the object (the package box in \textsl{"loading"}, the left person in \textsl{"judo"}) on each frame. The third row shows the proposals propagated to the key frame. \textbf{On the right side:} the first image is the voting of the object proposals inside a maximal clique on the {\graphnameshort}, which is created with all propagated object proposals initially generated on the key frame clip. The second image is the final binarized object mask we obtained. 
  }
  \vskip -0.15in
  \label{fig:motivation}
\end{figure*}



For robots to operate safely and reliably in dynamic environments or `in-the-wild', they must be able to discover novel unseen objects with no supervision from continuous video streams. Robots can be pre-trained to understand what general objects may look like, but once deployed in the field, it would be very difficult to supply them with additional annotations. It is especially likely that objects from novel categories unseen from the training set would need to be discovered. 
Hence, acquiring this capability of unsupervised object discovery from new videos, which more commonly is called Unsupervised Video Object Segmentation (UVOS)\cite{Caelles_arXiv_2019}, is an important research problem.

In the related problem of semi-supervised Video Object Segmentation (VOS), the first frame annotation is provided to the algorithm, which tracks and segments each object throughout the rest of the video. Most recent works typically utilize space-time transformers like STM \cite{Oh_2019_ICCV} which properly match the visual features in a new frame with previous frames using a deformable attention model. This helps the systems track objects across significant motion, deformation, and occlusion. 

Hence, a simple and natural idea to address the UVOS problem in prior work is to identify object proposals on a few key frames and then utilize a semi-supervised VOS algorithm to track them~\cite{luiten2020unovost}. Usually, instance segmentation algorithms such as Mask-RCNN~\cite{he2017mask} are utilized to identify object proposals in those key frames. However, for VOS to work well, the starting frame usually needs to be annotated with high precision, because wrongly annotated regions in this frame, serving as ground truth, could lead to significant drift in subsequent frames. Similarly, a missing part from the annotation might be missed forever because the tracker thinks it belongs to the background instead of the object. Hence, achieving high segmentation accuracy at those key frames is essential for better UVOS performance.


However, single-frame instance segmentation is often noisy and does not always provide the required precision in the key frames. In this paper, we present a novel approach to improve the segmentation on the key frames. The idea is to take into account object proposals from nearby frames and use them to jointly reason about the segmentation in the key frame, which allows segmentations from different frames to cancel out the noise in each other. 

Our approach builds a \textbf{M}ulti-frame \textbf{P}roposal \textbf{G}raph (MP-Graph) using object proposals initially generated in a local window around each key frame, and then locates maximal cliques in this graph from which the final segmentation on the key frame is generated. Each clique in the graph consists of multiple segments that correspond to the same object, hence jointly reasoning among all of them may generate more precise segmentations. Fig.~\ref{fig:motivation} shows two examples, where the segmentation of the object is poor on the key frame, meanwhile, none of the segmentations from the five frames are perfect. However, their joint voting produces a segmentation very close to the ground truth. Once better key frame segmentation is obtained, we can use any VOS algorithm to propagate it to the entire video and use a sequence non-maximum suppression (NMS) approach to filter out redundant objects. Our approach is lightweight and fast, and thus adds little computational overhead to the VOS algorithms used for tracking key frame proposals. Although our approach does not handle instance classification, we also extend our approach to a similar problem of Video Instance Segmentation (VIS), where it is required to classify the tracked object instances to a known set of categories, extending image instance segmentation to the video domain. 

We validate our approach through extensive experiments providing quantitative and qualitative analysis on both tasks. Experiments on the DAVIS-UVOS and Youtube-VIS benchmark show that the better key frame segmentations from our approach lead to state-of-the-art performance. Notably, our approach outperforms the state-of-the-art ~\cite{Lin_2021_ICCV} in UVOS that jointly trains the proposal generation model and the STM model. Not needing joint training is a significant advantage of our model -- this makes it future-proof because it can be then plugged in seamlessly to any future VOS models that achieve better performance without a cumbersome re-training process.

To summarize, our main contributions are,
\begin{enumerate}
\vspace{-0.07in}
    \item We propose \textbf{M}aximal \textbf{C}liques on \textbf{M}ulti-Frame \textbf{P}roposal
    \textbf{G}raph ({\methodname}), which utilizes maximal cliques over a graph of object proposals from a local window. Reasoning over the multiple similar proposals within these maximal cliques over this graph yields better object proposals. \methodname~
    is \textit{modular}, \textit{lightweight}, and \textit{fast}, enabling it to be plugged into any VOS algorithms that track object proposals without requiring the joint training that previous methods need.
\vspace{-0.06in}
    \item \methodname~ outperforms all SOTA unsupervised methods on the DAVIS-UVOS validation and test-dev set. Furthermore, it significantly improves the performance given the same single-image instance segmentations in the Video Instance Segmentation Task on the Youtube-VIS 2019 validation set benchmark.
\end{enumerate}

\section{Related Work}
\textbf{Image Instance Segmentation.} The task is to produce pixel-level predictions for each object instance in a frame. Top-down \cite{he2017mask,liu2018path,huang2019mask} approaches like Mask-RCNN \cite{he2017mask} and its follow-ups adopt the `detect-then-segment' paradigm. These two-stage approaches are accurate but relatively slow due to the exhaustive search process.

To overcome these drawbacks, bottom-up methods \cite{newell2016associative,de2017semantic,liu2017sgn} view the problem as `label-then-cluster' where the model learns an affinity function to group pixel embeddings belonging to the same object instance. 
Single-stage algorithms ~\cite{wang2020solo, wang2020solov2,yuan2020deep,xie2020polarmask} simplify computational-heavy post-processing, and in particular, SOLO \cite{wang2020solo} and SOLOv2 \cite{wang2020solov2} segment the object instances by
locations without using bounding boxes or metric learning. DETR \cite{carion2020end} inspires End-to-End transformer-based models \cite{dong2021solq, wang2021max, cheng2021per, cheng2022masked, thawakar2022video}. The most recent Mask2Former \cite{cheng2022masked} uses masked attention to achieve state-of-the-art performance in the instance segmentation task. 
 
\textbf{Semi-Supervised Video Object Segmentation.}
 Video Object Segmentation (VOS) can be applied to acquire pixel-level segmentations of primary objects in the scene given unconstrained videos. Depending on the level of supervision, they can be categorized as semi-supervised (one-shot), interactive, and unsupervised (zero-shot).
 Early work \cite{Cae17,perazzi2017learning,cheng2017segflow} fine-tuned a pretrained network at test-time using multiple data augmentations on the mask of each object from the first frame. They are usually very slow due to the excessive test-time fine-tuning. Their performance under occlusion and appearance changes is also limited due to the overfitting to the appearance of the first frame. Later approaches improved speed and accuracy through metric learning \cite{voigtlaender2019feelvos,chen2018blazingly}, guided propagation \cite{oh2018fast,Oh_2019_ICCV,yang2018efficient} and transformer-type networks \cite{Oh_2019_ICCV,li2020fast,nguyen2021space,wu2020memory,seong2020kernelized, cheng2021rethinking}. 


\textbf{Unsupervised Video Object Segmentation.} Early work utilizes motion patterns such as clustering object motion trajectories \cite{brox2010object,xie2019object,fragkiadaki2012video} or deep CNN-based spatio-temporal grouping \cite{dave2019towards,xie2019object}. Some combine appearance with optical flow for enhanced object features \cite{zhou2020motion,lu2019see,cheng2017segflow}, or use optical flow alone \cite{tokmakov2017learning}. 
A common drawback to these methods lies in their inability to be generalized to videos that have static objects, large motion blur, or cluttered backgrounds. 
 
 For multi-object VOS, learning appearance models of all the object proposals have been previously explored \cite{li2013video,wu2015robust}.
Currently, `track-by-detect' \cite{garg2021mask,luiten2020unovost,luiten2020unovost,Wang_2019_ICCV,Ventura_2019_CVPR} paradigm is popular where instance segmentation framework generates object proposals via Mask-RCNN \cite{he2017mask} which then are tracked consistently through a video sequence. UnOVOST \cite{luiten2020unovost} pruned tracklets from proposals into long-term tracks via visual similarity. In AGNN \cite{Wang_2019_ICCV}, mask proposals over a video sequence were aggregated via graph neural networks. Most recently, \cite{Zhou_2021_CVPR} proposed a novel instance segmentation, tracking, and re-identification network.

\textbf{Video Instance Segmentation.}
The VIS task was proposed in MaskTrack R-CNN \cite{yang2019video} which adds a tracking head to Mask RCNN and an external memory to store and associate features
of object instances across multiple frames.
This tracking paradigm is extended in \cite{bertasius2020classifying, cao2020sipmask} and STEMseg \cite{athar2020stem} models video clips as 3D space-time volumes to predict masks by clustering learned embeddings. An application of graph neural networks is seen in VisSTG~\cite{wang2021endGNN}. 
Propose-Reduce \cite{Lin_2021_ICCV} generates instance sequence proposals on key frames and reduces redundant sequences of the same instances with non-maximum suppression. Transformer-based techniques have become increasingly successful \cite{wang2021end, hwang2021video, wu2022seqformer, mssts2022} applying cross-attention to process video clips. Mask2Former is extended to VIS \cite{cheng2022masked} by directly making predictions on the entire video sequence. Online VIS methods also exist, but they usually have lower accuracy due to not observing the entire sequence \cite{han2022visolo,wu2022efficient}.


\section{Proposed Method}
\begin{figure*}[ht]
\centering
\includegraphics[width=\linewidth]{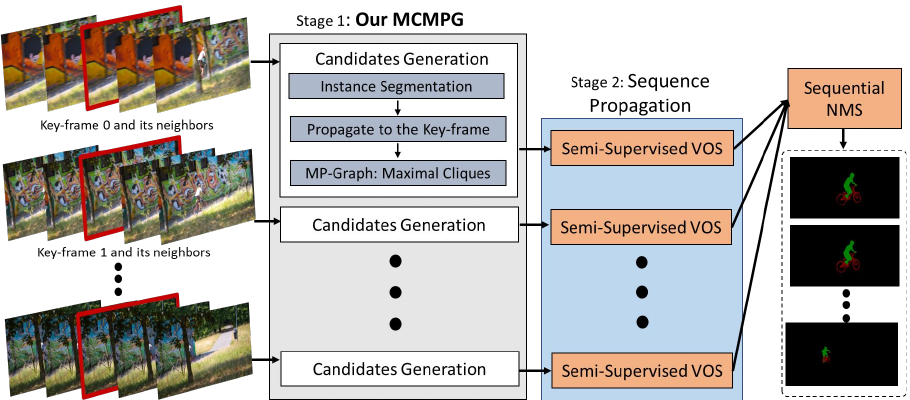}
\caption{Architecture of the proposed \methodname~ algorithm for the UVOS task. It includes 2 stages: (1) \methodname: key frame proposals are generated from each \textit{key frame clip}, including the key frame and its neighboring frames; (2) a tracker tracks these proposals bidirectionally through the whole sequence to obtain the final unsupervised video object segmentation. Our main contribution \methodname~ includes 3 steps: 1) Run instance segmentation on each frame in the key frame clip; 2) Propagate the proposals to the key frame; 3) Create the {\graphnameshort} from the propagated proposals and locate its maximal cliques. By combining proposals within these maximal cliques, we can obtain object proposals that are better than the instance segmentation result of any frame and subsequently improve the performance of the UVOS task. (Best viewed in color)}
\label{fig:main-pipeline}
\vskip -0.15in
\end{figure*}
The architecture of using \methodname~ to perform UVOS is shown in Fig. \ref{fig:main-pipeline}. \methodname~ aims to generate the key frame proposals with higher quality by creating a {\graphnameFullsmall} and finding its maximal cliques. Afterward, any semi-supervised VOS algorithm can be used to track each instance proposal to the beginning and end of the sequence. 
Lastly, we adopt sequence NMS in order to remove duplicate segments. 

\subsection{Problem Definition}
Given a set of RGB frames $\mathcal{I}= \{I_t\}^{T-1}_ {t=0} $ where $I_t \in \mathcal{R}^ {3\times h \times w} $ and $T$ is the total number of frames, the goal is to produce a sequence of consistent segmentation masks $\mathcal{S}= \{M_t\}^{T-1}_ {t=0} $ for each of the $m$ objects in the video. where $M_t \in \mathcal{R}^ {m\times h \times w} $ represents the masks for all of the objects.

\subsection{Proposal Generation on Key Frames}
As we argued in the introduction, the quality of the generated key frame proposals is crucial for successful unsupervised video object segmentation. However, as Fig.~\ref{fig:motivation} shows, even state-of-the-art instance segmentation approaches can generate bad segments in some frames due to motion blur, occlusion, and object poses that are very different from the training set. In this section, we introduce a new algorithm that improves key frame proposals by joint reasoning from many frames. The algorithm is illustrated in Alg.~\ref{algo:proposal}. 

\textbf{Key Frame Selection.} 
Similar to \cite {Lin_2021_ICCV}, we select $K$ key frames with fixed intervals. Namely, for a $T$-frame video, $K$ frames $\{I_ {g (0), \cdots, I_{g(K-1)}} \} $ are selected evenly starting from frame $0$. 
\begin{equation}
        g(k) = k  \max(\lfloor T/K \rfloor), 1), k=0, \cdots, K-1
\end{equation}

In contrast to \cite{Lin_2021_ICCV} that uses the segments in the key frames directly as key frame proposals to track, we select a key frame clip to generate the key frame proposals. 
Here, the key frame clip on a key frame $I_{g(k)}$ is $I^C_{g(k)} = \{I_i | i= g(k)- \frac{H-1}{2}, \cdots, g(k)+\frac{H-1}{2} \}$. It contains the key frame itself and its $H-1$ neighbors from a local window centered around the key frame. 

\textbf{Instance Segmentation and Propagation.} With the key frame clip $I^C_{g(k)} $, we first run image-level instance segmentation in each frame individually. Then we propagate all the segments $S^I_i$ to the key frame $g(k)$ by using the same tracking algorithm that we use in the later stage (Stage 2 in Fig.~\ref{fig:main-pipeline}). The set of propagated proposals is denoted as $S^{g(k)} = \{S^{g(k)}_i | i=g(k)-\frac{H-1}{2}, \cdots, g(k) + \frac{H-1}{2} \} $, where $S^{g(k)}_i \in R^{L_i\times h \times w}$ is the probability masks of $L_i$ proposals that are segmented in frame $i$ and then propagated to the key frame $g(k)$. This set can be used to create the {\graphnameFullsmall} which is introduced below.
 
\textbf{{\graphnameshort} ({\graphnameFullcapital}).} On the proposal set $S^{g(k)} $ from one key frame clip, we create an undirected graph where 
    each propagated proposal in $S^{g(k)}$ is one vertex in the graph, and an edge is created between a pair of vertices if their Intersection-over-Union (IoU), Eq.(\ref{eq::iou})) is larger than $t_0$.
\begin{equation}
    IoU(S^{g(k)}_{i, o_1}, S^{g(k)}_{j,o_2}) = \frac{S^{g(k)}_{i, o_1} \cap S^{g(k)}_{j,o_2}}{S^{g(k)}_{i, o_1} \cup S^{g(k)}_{j,o_2}}
    \label{eq::iou}
\end{equation}
where $S^{g(k)}_{i, o_1}$, $S^{g(k)}_{j,o_2}$ are two propagated proposals from the temporal frame $i$ and $j$ to the key frame $g(k)$ so that their IoU is measured in the same key frame.

\begin{figure}[ht]
  \centering
  \includegraphics[width=\linewidth]{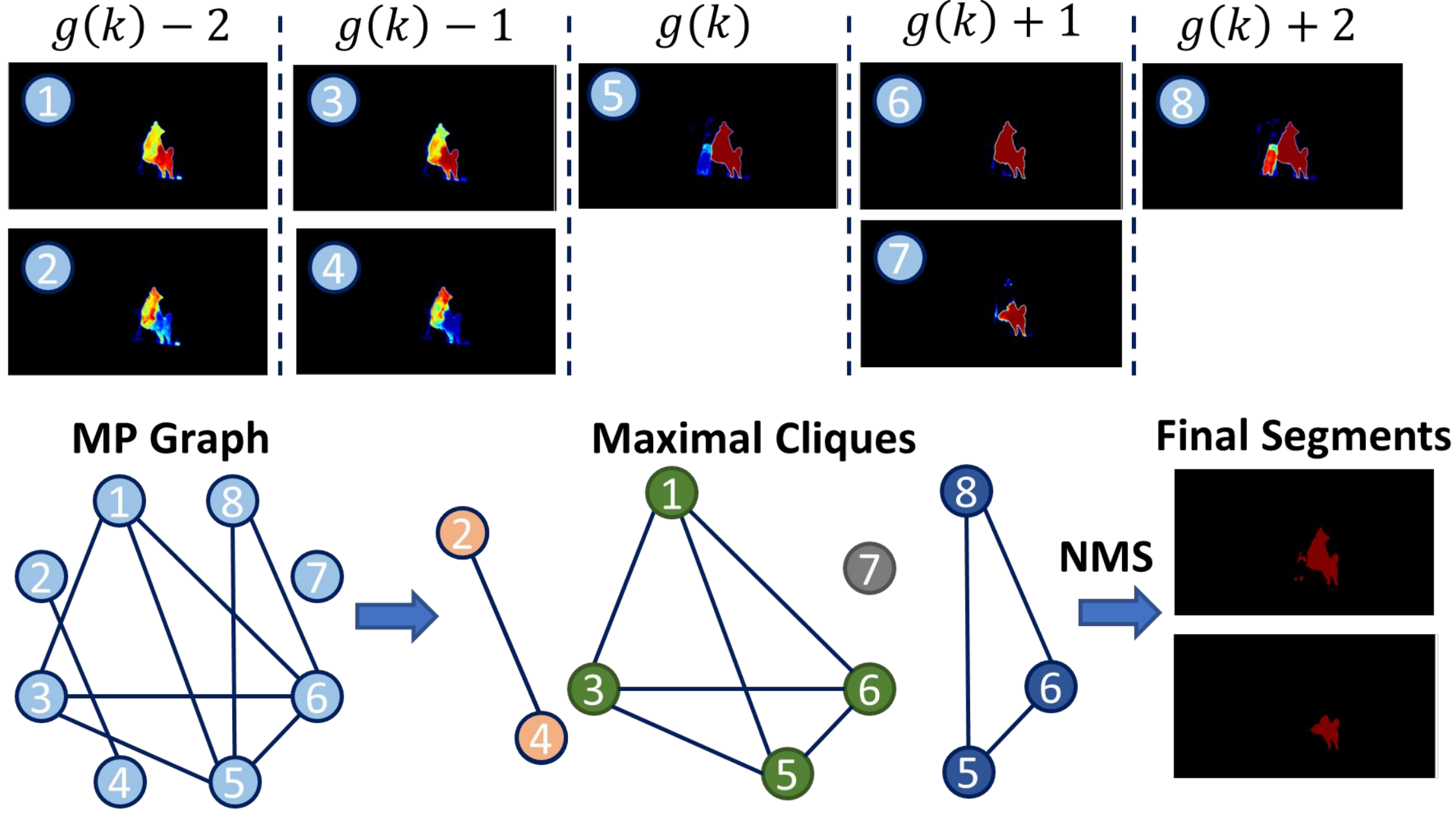}
  \caption{An example of Multi-frame Proposal Graph. Propagated segments are connected based on their IoU on the key frame, then one segment is generated from each clique (Best viewed in color)}
  \label{fig:maximal_clique}
  \vskip -0.1in
\end{figure}

\begin{algorithm}[htb]
    \SetKwFunction{isOddNumber}{isOddNumber}
    \SetKwInOut{KwIn}{Input}
    \SetKwInOut{KwOut}{Output}

    \KwIn{key frame and its neighbours \{$I_i | i = {g(k)-\frac{H}{2}}, \cdots, I_{g(k)+\frac{H}{2}}\}$}
    \KwOut{Instance Proposals S in the key frame}
    
    \For{$i \leftarrow g(k)-\frac{H}{2}$ \KwTo $g(k) + \frac{H}{2}$}{
        $S^I_i \leftarrow InstanceSegmentation(I_i)$
        $S^{g(k)}_i \leftarrow Propagate(S_i, \{I_i \cdots I_{g(k)}\})$
    }
    $G = {\graphnameshort}(S^{g(k)})$  
    $Cliques = G.maximalCliques()$

    \For{$C \in Cliques$}{
        $S^c \leftarrow combine(C, S^{(g(k))})$ \tcp*[f]{Eq.\ref{Eq:clique-merge}}
    }
    $S_k \leftarrow \cup S^c$
    
    \KwRet{$S_k$}
    \caption{Key frame Proposal Generation}
    \label{algo:proposal}
\end{algorithm}
We name the graph as the {\graphnameFullcapital} given that its nodes are propagated proposals from different temporal frames and its edges are created based on the spatial IoU computed in the same time frame $g(k)$. After the {\graphnameshort} is created, we adopt the \textit{maximal clique algorithm}~\cite{bron1973algorithm} to generate the final instance proposals in the key frame to be tracked (See Fig. \ref{fig:maximal_clique}). 

\textbf{Key Frame Proposals.} 
In an undirected graph, a \textit{clique} is a complete sub-graph in which every two vertices are adjacent. A \textit{maximal clique} is a clique that \textit{cannot be extended} by including \textit{any more adjacent} vertex. The largest maximal clique is called a \textit{maximum clique}. Accordingly, in the {\graphnameshort}, a \textit{maximal clique} is a subset of propagated proposals which all significantly overlap each other.

Given a maximal clique $C$ that contains $n$ propagated proposals $\{O_i | i=0, \cdots, n-1\}, n \leq H$, its corresponding key frame object proposal $S^C$ is computed as: 
\begin{equation} 
    S^C = (\frac {1}{H}\sum_{i=0,\cdots n-1}O_i) \geq t_1 
    \label{Eq:clique-merge} 
\end{equation} 
where $t_1$ is a threshold that can be set to a small value ($0.2$ in the experiments) without introducing noise in the segmentation. In the \textsl{"loading"} example in Fig.\ref{fig:motivation}, the propagated proposals from $g(k)-2, g(k)$, and $g(k)+2$ have low confidence on the bottom left part of the package box object and high confidence on the right corner, while the propagated proposal from $g(k-2)$ has high confidence on the bottom left part and has low confidence on the right corner. Thus, the maximal clique including the proposals propagated from $g(k)-2, g(k)-1, g(k)$, and $g(k)+2$ can smooth the score over the object and then segment the package correctly.  
Notably, the algorithm can retrieve the object proposals even when the segmentation on the key frame is poor. In the \textsl{"judo"} example presented in Fig.\ref{fig:motivation}, the segments of the person on the left are noisy in the first three frames due to serious motion blur, but the final, combined proposal for the person in the clique consisting of propagated proposals from frame $g(k)-2$, $g(k)+1$, and $g(k)+2$, is a better segmentation result due to predictions in the last two frames that fill in the missing pixels and remove the false pixels. 

\subsection{Tracking}

\textbf{Sequence propagation.} After the key frame proposals are obtained, an off-the-shelf semi-supervised VOS method can be used to track all the objects from the key frame bidirectionally through the video sequence. In this way, we can always plug in state-of-the-art VOS algorithms for better performance.
The propagation results from each key frame $\{S_k | k =0, \cdots, K-1\}$ are then concatenated together as $\hat{S} \in \mathcal{R}^{T \times N \times h \times w}$, assuming we segment $N$ objects in total after the tracking stage.

\textbf{Sequence Score.} We compute the score of an object sequence $\hat{S}_o$ by referring to the object masks $S^I$ and objectness scores $C^I$ obtained from the instance segmentation model. The sequence score is then computed by 
\begin{equation} \label{eq:seq_score}
    Score_o = \frac{1}{T}\sum_{t = 0, \cdots, T-1} \max_i (IoU(\hat{S}_o^t, S^I_i)C^I_i).
\end{equation}
where $i$ iterates over the object proposals from the instance segmentation on frame $t$. We utilize the objectness scores $C^I_i$ from the instance segmentation model without re-computing the objectness of the new object masks coming from tracking. This avoids re-running the objectness computation for the merged mask. 
The sequence score for each object sequence is used in Sequence NMS for removing duplicate object sequences that are detected in different key frames. It is also used to rank object sequences in the case where the algorithm is allowed to output only a fixed number of detections. Our sequence NMS follows \cite{Lin_2021_ICCV} which removes overlapping tracks by running the traditional NMS algorithm with the tracking scores and the sequence IoU.

\section{Experiments}
\subsection{Implementation Details}
\textbf{Architecture.} Our proposed architecture is shown in Fig.~\ref{fig:main-pipeline}. $K$ key frames are sampled from a given video sequence and the corresponding key frame clips are fed to a network to obtain image-level object segmentation proposals. Then the second network is utilized to propagate the initial segments to their related key frames. Afterwards, the {\graphnameshort} is built to compute the maximal cliques. Each maximal clique generates one key frame proposal by averaging over the propagated proposals in the clique and then binarizing with $t_1$. The key frame proposals with areas smaller than $10$ pixels are discarded. Finally, the second network is used again to track these key frame proposals to the beginning and end of the video.

This architecture is modular, so it is easy to replace both the instance segmentation and tracking methods with state-of-the-art algorithms. The methods we adopted for DAVIS and Youtube-VIS are discussed in Sec. \ref{sec:sub:davis} and Sec. \ref{sec:sub:ytvis} respectively. 

\textbf{Training / Inference.} With pre-trained models for instance segmentation and semi-supervised VOS network, our approach does not need extra training. In our experiments (except ablations), the number of key frames $K$ is set to $2$ on DAVIS and $8$ on Youtube-VIS, the size of key frame clip $H = 3$, threshold $t_0 = 0.5, t_1=0.2$.




\begin{table*}[!htp]
    \centering
    \caption{Quantitative video multi-object segmentation results on DAVIS 2017 \textsl{val}.}
    \vskip -0.1in
    \resizebox{0.95\textwidth}{!}{%
    \begin{tabular}{|l|c|c|c|c|c|c|c|c|c|}
        \hline
        \multicolumn{1}{|l|}{Methods} & Instance Seg. & backbone & $\mathcal{J}$ \& $\mathcal{F}$ Mean & $\mathcal{J}$ -Mean & \begin{tabular}[c]{@{}c@{}}$\mathcal{J}$ - Recall\end{tabular} & \begin{tabular}[c]{@{}c@{}}$\mathcal{J}$ - Decay\end{tabular} & $\mathcal{F}$ -Mean & \begin{tabular}[c]{@{}c@{}}$\mathcal{F}$ - Recall\end{tabular} & \begin{tabular}[c]{@{}c@{}}$\mathcal{F}$ - Decay\end{tabular} \\ \hline
        RVOS \cite{Ventura_2019_CVPR} & - & ResNet-101 & 41.2 & 36.8 & 40.2 & 0.5 & 45.7 & 46.4 & 1.7 \\
        PDB \cite{song2018pyramid} & - & ResNet-50 & 55.1 & 53.2 & 58.9 & 4.9 & 57.0 & 60.2 & 6.8 \\
        AGS \cite{wang2020paying} & - & ResNet-101 & 57.5 & 55.5 & 61.6 & 7.0 & 59.5 & 62.8 & 9.0 \\
        ALBA \cite{goel2018unsupervised} & - & ResNet-101  & 58.4 & 56.6 & 63.4 & 7.7 & 60.2 & 63.1 & 7.9 \\
        MATNet \cite{zhou2020motion} & - & ResNet-101 & 58.6 & 56.7 & 65.2 & -3.6 & 60.4 & 68.2 & 1.8 \\
        AGNN \cite{Wang_2019_ICCV} & - & ResNet-101 & 61.1 & 58.9 & 65.7 & 11.7 & 63.2 & 67.1 & 1.2 \\
        STEm-Seg  \cite{athar2020stem} & - & ResNet-101 & 64.7 & 61.5 & 70.4 & \textbf{-4.0} & 67.8 & 75.5 & 1.2 \\
        UnOVOST \cite{luiten2020unovost} & - & ResNet-101 & 67.9 & 66.4 & 76.4 & -0.2 & 69.3 & 76.9 & \textbf{0.0} \\
        Target-Aware \cite{Zhou_2021_CVPR} & - & ResNet-101  & 65.0 &  63.7 & 71.9 & 6.9 & 66.2 & 73.1 & 9.4 \\  
        Propose-Reduce \cite{Lin_2021_ICCV} & Mask-RCNN & ResNet-101  & 68.3 &  65.0 & - & - & 71.6 & - & - \\
        Propose-Reduce \cite{Lin_2021_ICCV} & Mask-RCNN & ResNeXt-101  & 70.6 &  67.2 & - & - & 73.9 & - & - \\
         \hline
         \methodname~ + STM ($w/o$ \graphnameshort) & Mask-RCNN & ResNeXt-101 &  70.0 & 67.1 & 73.0 & -1.1 & 72.3 & 80.0 & 0.9 \\ 
        \methodname~ + STM ($w/$ \graphnameshort)& Mask-RCNN & ResNeXt-101 &  71.7 & 68.9 & 74.6 & -4.9 & 75.8 & 83.2 & -2.1 \\ 
        \hline
        \methodname~ + STCN ($w/o$ \graphnameshort) & Mask-RCNN & ResNet-101 &  73.6 & 70.2 & 77.5 & -2.3 & 77.1 & 83.4 & 0.2 \\ 
        \methodname~ + STCN ($w/$ \graphnameshort)& Mask-RCNN & ResNet-101 &  76.8 & 73.8 & 81.9 & -1.2 & 79.2 & 85.5 & 1.9 \\ 
        \hline
         \methodname~ + STM ($w/o$ \graphnameshort)& SOLOv2 & ResNet-101 &  71.2 & 68.2 & 76.5 & -2.2 & 74.0 & 81.2 & 0.9 \\
        \methodname~+ STM ($w/$ \graphnameshort)& SOLOv2 & ResNet-101 &  72.5 & 69.0 & 77.3 & -3 & 76.1 & 83.3 & 5.3 \\ 
        \hline
        \methodname~+ STM ($w/o$ \graphnameshort)& SOLOv2 & ResNeXt-101 &  72.7 & 69.9 & 76.6 & -3.7 & 75.5 & 82.7 & -1.1 \\ 
        \methodname~+ STM ($w/$ \graphnameshort)& SOLOv2 & ResNeXt-101 &  \textbf{78.4} & \textbf{75.4} & \textbf{83.9} & 0.05 & \textbf{81.4} & \textbf{88.9} & 0.04 \\
        \hline
    \end{tabular}}
    \label{tab:UVOS}
\end{table*}

\begin{table*}[!htp]
    \vskip -0.1in
    \centering
    \caption{Quantitative video multi-object segmentation results on DAVIS 2019
    \textit{test-dev}.}
    \vskip -0.1in
    \resizebox{0.91\textwidth}{!}{%
    \begin{tabular}{|l| c | c|c|c|c|c|c|c|}
        \hline
        \multicolumn{1}{|l|}{Methods} & Backbone & $\mathcal{J}$ \& $\mathcal{F}$ Mean & $\mathcal{J}$ -Mean & \begin{tabular}[c]{@{}c@{}}$\mathcal{J}$ - Recall\end{tabular} & \begin{tabular}[c]{@{}c@{}}$\mathcal{J}$ - Decay\end{tabular} & $\mathcal{F}$ -Mean & \begin{tabular}[c]{@{}c@{}}$\mathcal{F}$ - Recall\end{tabular} & \begin{tabular}[c]{@{}c@{}}$\mathcal{F}$ - Decay\end{tabular} \\ 
        \hline
        RVOS \cite{Ventura_2019_CVPR} & ResNet-101 & 22.5 & 17.7 & 16.2 & 1.6 & 27.3 & 24.8 & 1.8 \\
        PDB \cite{song2018pyramid} & ResNet-50 & 40.4 & 37.7 & 42.6 & 4.0 & 43.0 & 44.6 & 3.7 \\
        AGS \cite{wang2020paying} & ResNet-101 & 45.6 & 42.1 & 48.5 & 2.6 & 49.0 & 51.5 & 2.6 \\
        UnOVOST \cite{luiten2020unovost}& ResNet-101 &58.0 & 54.0 & 62.9 & 3.5 & 62.0 & 66.6 & 6.6 \\
        Target-Aware \cite{Zhou_2021_CVPR} & ResNet-101 & 59.8 & 56.0 & \textbf{65.1} & 7.8 & 63.7 & 68.4 & 11.0 \\ \hline
        \methodname~ + STM ($w/${\graphnameshort}) & ResNet-101 &\textbf{61.2} & \textbf{56.1} & 63.5 & \textbf{-0.2} & \textbf{66.4} & \textbf{71.9} & \textbf{-0.5} \\ \hline
    \end{tabular}}
    \label{tab:UVOS-test}
    \vskip -0.1in
\end{table*}
\subsection{Unsupervised Video Object Segmentation}\label{sec:sub:davis}
\textbf{Dataset.} The DAVIS 2017 \cite{pont20172017} benchmark is used for video multi-object segmentation with high-quality masks for salient objects. It consists of 60 sequences used for training and 30 for validation. DAVIS 2019 \cite{Caelles_arXiv_2019} is an extension of DAVIS 2017 for the UVOS task. It has the same training and validation set as DAVIS 2017 and 30 new sequences in its \textsl{test-dev} set. To demonstrate that our proposed {\graphnameshort} is network-agnostic and can work with a wide range of instance segmentation models, we show experimental results with both SOLOv2 \cite{wang2020solov2} and Mask-RCNN instance segmentations from \cite{Lin_2021_ICCV}. SOLOv2 model is initialized with COCO pretrained weights. We then finetune its kernel branch and feature branch for 10 epochs, then the FPN for 5 epochs, and finally the ResNet-101 backbone blocks from the last block to the first block for 5 epochs per block. Similarly, for the tracker, we show experimental results with both STM \cite{Oh_2019_ICCV} and STCN\cite{cheng2021rethinking}. 


\textbf{Metric.} We follow the standard evaluation settings \cite{perazzi2016benchmark}: the performance is reported in terms of region similarity $\mathcal{J}$ , boundary accuracy $\mathcal{F}$, and the overall metric $\mathcal{J} \& \mathcal{F}$. The
evaluation scores on the \textsl{test-dev} set are obtained from
the evaluation server of the DAVIS 2019 challenge.
 
\textbf{Results on DAVIS 2017 \textsl{val}.} In Table \ref{tab:UVOS}, we compare our approach with state-of-the-art unsupervised video multi-object segmentation methods on the DAVIS 2017 dataset. The common baselines from published works are included: RVOS \cite{Ventura_2019_CVPR}, PDB \cite{song2018pyramid}, AGS \cite{wang2020paying}, ALBA \cite{wang2020paying}, MATNet \cite{zhou2020motion}, AGNN \cite{Wang_2019_ICCV}, and Stem-Seg \cite{athar2020stem}. Some of the recent top-ranked methods include UnOVOST \cite{luiten2020unovost}, Target-Aware \cite{Zhou_2021_CVPR}, and 
Propose-Reduce \cite{Lin_2021_ICCV}. 
As shown in Table \ref{tab:UVOS}, on DAVIS 2017 \textit{val}, our approach achieves the highest overall results across most metrics. Prior methods such as UnOVOST and MATNet are computationally expensive and also need to compute optical flow for motion estimation. Our work requires only RGB frames as input and outperforms the previous best method Propose-Reduce \cite{Lin_2021_ICCV} by 1.7\% in terms of $\mathcal{J}$ \& $\mathcal{F}$-Mean when using the same instance segmentation method and the same ResNeXt-101 backbone. With the more complex ResNeXt-101 backbone, we outperform \cite{Lin_2021_ICCV} by $7.8\%$. 
Note that by utilizing frames next to the key frames, our approach may be thought of as utilizing more frames than \cite{Lin_2021_ICCV}. However, the ablation study in \cite{Lin_2021_ICCV} shows that more key frames do not further help their performance on this dataset, which shows the importance of \methodname~in terms of combining and refining the proposals.

In Table \ref{tab:UVOS}, we also adopt the Mask-RCNN module from \cite{Lin_2021_ICCV} to compute the key frame proposals. This approach without the \graphnameshort~ achieves a score $1.7\%$ higher than \cite{Lin_2021_ICCV} since we adopt a separate STM model to perform tracking. Adding {\graphnameshort} improves another $1.7\%$ over this baseline, which shows the effectiveness of \methodname~even with the same object proposal algorithm as \cite{Lin_2021_ICCV}. The modular design of the proposed framework also makes it easy for a different tracking algorithm to be plugged in without joint training. In order to show this, we report the performance of \methodname~with STCN \cite{cheng2021rethinking} as well.

\textbf{Results on DAVIS 2019 \textsl{test-dev}.} We evaluate the proposed approach \methodname~on DAVIS 2019 \textsl{test-dev} set shown in Table \ref{tab:UVOS-test}. Our approach achieves the state-of-the-art result in terms of $\mathcal{J}$ \& $\mathcal{F}$-Mean at $61.2$. Compared with the previous state-of-the-art Target-Aware \cite{Zhou_2021_CVPR}, our approach improves significantly on the boundary F-metric, which shows that our proposals cover object boundaries significantly better. Here we do not test the ResNeXt-101 backbone for a fair comparison with prior work, which also does not use this more complex backbone.

\subsection{Video Instance Segmentation}\label{sec:sub:ytvis}

\textbf{Video Instance Segmentation (VIS).} Different from UVOS which segments salient object instances, VIS aims at discovering and segmenting all object instances of pre-defined object categories from videos. It requires predictions for both object segmentations and object categories. Usual VIS approaches contain a category classification head to predict the category score.
 
 We adapt {\methodname} to the VIS domain by adopting the Mask-RCNN module from \cite{Lin_2021_ICCV} to generate object segments and the category scores on each frame as different settings. Meanwhile, STM \cite{Oh_2019_ICCV} is used to propagate object segments generated on frames in a key frame clip to the key frame and to track key frame proposals bidirectionally throughout the videos. We also test our approach to the task by utilizing the latest transformer-based method, Mask2Former-VIS from \cite{cheng2022masked}, as the instance segmentation network.

\textbf{Dataset.} YouTube-VIS 2019 \cite{yang2019video} is a large-scale dataset for VIS with objects in multiple categories. It contains $2,283$ high-resolution YouTube videos for training and $302$ for validation, covering 4,883 unique object instances out of $40$ categories. We use this dataset to examine the performance of our model in more challenging scenarios.

\textbf{Metrics.} YouTube-VIS adopted the standard evaluation metrics in image instance segmentation, average precision (AP) and average recall (AR), to evaluate performance. It follows COCO evaluation \cite{lin2014microsoft} to compute AP by averaging it over multiple intersection-over-union (IoU) thresholds from 50\% to 95\% at step 5\%.

\textbf{Results on YouTube-VIS 2019 \textsl{val.}} We compare our approach with state-of-the-art video VIS approaches on the YouTube-VIS 2019 benchmark. As shown in Table \ref{tab:VIS-2019}, our approach achieves consistent improvements with all different backbones and instance segmentation methods. 
Specifically, adding the {\graphnameshort} achieves at least 1.0\% higher over the baseline without the {\graphnameshort}, which shows the effectiveness of the {\methodname}. Note that the results on the YouTube-VIS are significantly affected by the accuracy of the object categorization, which is orthogonal to our contribution to improving the key frame segmentation. Hence, our AP@50 is not necessarily the best, since this metric is mainly affected by classification accuracy, but our higher AP and higher AP@75 indicate better segmentation quality 
our approach achieves. Compared with SeqFormer~\cite{wu2022seqformer}, our AP is higher, but AP@50 and AP@75 are both slightly lower. This shows that we very likely have achieved significantly better performance in the AP regimes even higher than $75\%$ IoU, greatly indicating the strong segmentation quality our approach provides.

\begin{table*}[!htp]
    \centering
    \caption{Results on YouTube-VIS 2019 \textsl{val}.}
    \vskip -0.1in
    \resizebox{0.92\textwidth}{!}{
    \begin{tabular}{|l|c|c|c|c c c c|}
        \hline
        \multicolumn{1}{|l|}{Methods} & Instance Seg. & backbone & \hspace{0.2cm}AP\hspace{0.2cm} & AP@50 & AP@75 & AR@1 & AR@10  \\ \hline
        SipMask \cite{cao2020sipmask} & - & ResNet-50 & 33.7 & 54.1 & 35.8 & 35.4 & 40.1 \\
        STEm-Seg  \cite{athar2020stem} & - & ResNet-101 & 34.6 & 55.8 & 37.9 & 34.4 & 41.6 \\
        Target-Aware \cite{Zhou_2021_CVPR} & - & ResNet-101  & 37.1 &  57.1 & 40.9 & 34.8 & 43.2 \\  
        Propose-Reduce \cite{Lin_2021_ICCV} & - & ResNet-101  &  43.8 & 65.5 & 47.4 & 43.0 & 53.2 \\ 
        Propose-Reduce \cite{Lin_2021_ICCV} & - & ResNeXt-101  &  47.6 & \textbf{71.6} & 51.8 & 46.3 & 56.0 \\ 
        \hline
        {\methodname} ($w/o$ \graphnameshort) & Mask-RCNN & ResNet-101 &  43.4 & 64.4 & 48.9 & 45.0 & 57.1 \\
        {\methodname} ($w$ \graphnameshort)  & Mask-RCNN & ResNet-101 &  44.6 & 64.2 & 49.5 & 46.4 & 58.5 \\ 
        \hline
        {\methodname} ($w/o$ \graphnameshort) & Mask-RCNN & ResNeXt-101 &  47.4 & 70.6 & 52.3 & 47.5 & 60.0 \\
        {\methodname} ($w$ \graphnameshort) & Mask-RCNN & ResNeXt-101 &  \textbf{48.4} & 70.4 & \textbf{52.7} & \textbf{48.6} & \textbf{60.1} \\
        \hline
        \multicolumn{8}{|l|}{transformer-based methods} \\
        \hline
        SeqFormer\cite{wu2022seqformer} & - & ResNet-101 & 49.0 & 71.1 & \textbf{55.7} & \textbf{46.8} & \textbf{56.9} \\
        Mask2Former\cite{cheng2022masked} & Mask2Former-VIS & ResNet-101 & 49.2 & \textbf{72.8} & 54.2 & - & - \\
        \hline
        {\methodname}($w/o$ \graphnameshort) & Mask2Former-VIS & ResNet-101 &  48.5 & 65.7 & 53.0 & 43.7 & 54.0 \\
        {\methodname}($w$ \graphnameshort) & Mask2Former-VIS & ResNet-101 &  \textbf{50.5} & 70.3 & 55.0 & 45.0 & 55.8 \\ 
        \hline
    \end{tabular}}
    \label{tab:VIS-2019}
\end{table*}

\subsection{Ablation studies }  
\begin{table}[!htp]
    \centering
    \caption{Ablation Study on DAVIS 2017 \textsl{val}  on the influence of the number of key frames for with (w/) and without (w/o) using \graphnameshort~to generate refined object proposals.}
    \resizebox{0.47\textwidth}{!}{%
    \begin{tabular}{c || c | c }
    \hline
    No. of key frames & 
    \begin{tabular}[c]{@{}c@{}}$\mathcal{J}$ \& $\mathcal{F}$-Mean \\ w/o {\graphnameshort} \end{tabular}  &
    \begin{tabular}[c]{@{}c@{}} $\mathcal{J}$ \& $\mathcal{F}$-Mean \\ w/ {\graphnameshort} $H=3$ \end{tabular}  \\ \hline
    1 & 68.1 & 69.0 (+0.9) \\
    2 & 70.5 & 72.5 (+2.0) \\
    3 & 69.7 & 72.0 (+2.3) \\
    4 & 71.2 & 72.0 (+0.8) \\
    5 & 70.8 & 72.4 (+1.6) \\ \hline
    \end{tabular}
    }
    \label{tab:keyframe-num}
\end{table}

\textbf{Key Frame Selection.} 
In Table \ref{tab:keyframe-num}, we illustrate the effectiveness of \methodname~on combining and refining the proposals with different numbers of key frames in Table \ref{tab:keyframe-num} on DAVIS 2017 \textsl{val}. 
Without {\graphnameshort}, it reduces to the baseline approach similar to \cite{Lin_2021_ICCV}, and the results in the table are comparable to \cite{Lin_2021_ICCV} as well. 
With {\graphnameshort} ($H = 3$), better segments that are obtained from merging proposals in the same clique show to improve performance at every setting of key frames, which validates that the proposed approach provides significant performance improvement as discussed in the paper. Also, we observe that the performance is robust to different numbers of key frames and satisfactory with just $2$ key frames, not requiring an excessive amount of key frames to obtain good performance on this dataset.

\textbf{Object proposal quality with {\graphnameshort}.} 
 Here we provide a comparison of the object proposals quality with and without the {\graphnameshort} on the key frames evaluated against the ground truth objects on the DAVIS 2017 \textsl{val} set. {\graphnameshort} improves the key frame proposals by $3.7\%$ in terms of \textsl{mIoU} as shown in Table \ref{tab:davis-proposal-mIoU}. This significant improvement in the quality of the key frame proposal is the main driver of the performance of our approach.
 
\begin{table}[!htp]
    \caption{Quality of the key frame proposals on DAVIS 2017 \textsl{val}.}
    \centering
        \begin{tabular}{|c|c|c|}
            \hline
            & \textsl{w/} {\graphnameshort} & \textsl{w/o} {\graphnameshort} \\
            \hline 
            mIoU (\%)   & 79.2 & 75.5  \\
            \hline
        \end{tabular}
    \label{tab:davis-proposal-mIoU}
\end{table}

\subsection{Run-Time Analysis.}
\begin{figure}[!ht]
  \centering

  \includegraphics[width=\linewidth]{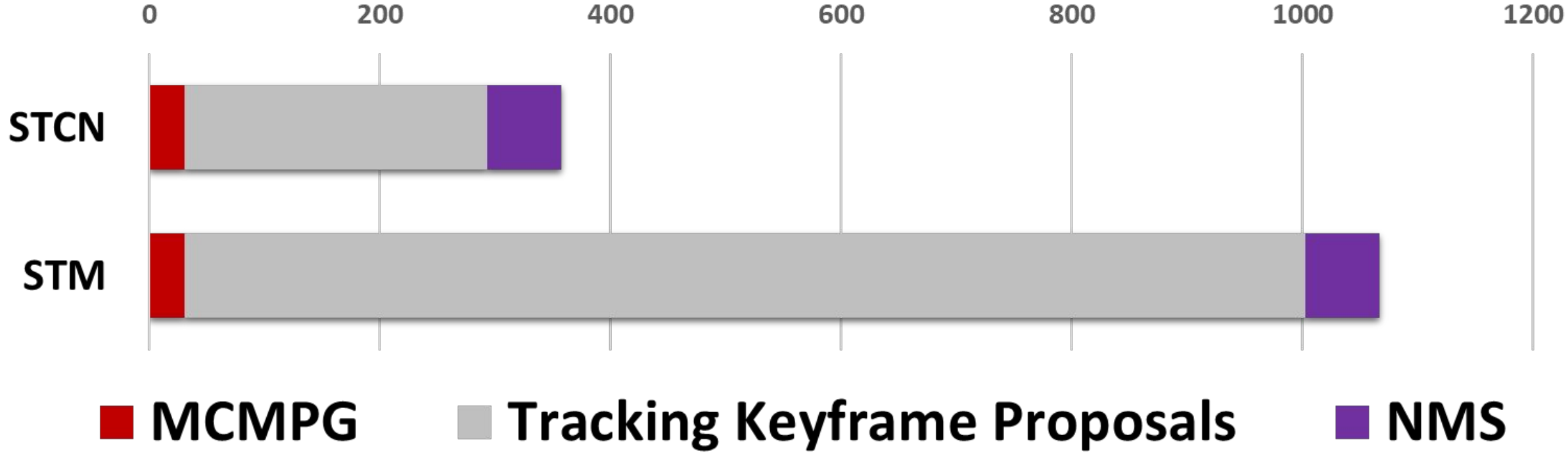}
  \caption{Run-time (in seconds) of {\methodname} on DAVIS-UVOS 2017 \textsl{val} with $2$ key frames. Note that MCMPG is fast (only $\mathbf{8.7}\%$ or $\mathbf{2.9}\%$ of the running time for STCN and STM VOS models, respectively), while improving the final tracking performance significantly.  The bottleneck of the speed comes from an external tracking algorithm such as STM which requires re-running the backbone network for each proposal. Alternatively, one could use a newer tracking algorithm such as STCN where all the proposals can share the same backbone features, which would make the system much faster.}
  \label{fig:run_time}
  \vskip -0.15in
\end{figure}

 We report the run-time of each module in \methodname~in Fig. \ref{fig:run_time}. The results are generated by using an NVIDIA Tesla V100 GPU. It shows that the process of generating proposals and improving them with the MP-Graph is very lightweight and takes minimal time to run. The most time-consuming part of the system is the semi-supervised VOS module. The limitation of {\methodname} is that the VOS module needs to track objects starting from multiple key frames. 
 In the STM algorithm, the encoder takes the object mask as \textit{input}. Hence, with each new object mask, the backbone has to be run again, which is quite suboptimal, especially for our approach which requires running tracking on a significantly larger amount of proposals than the regular semi-supervised VOS task for which STM was designed. 
 
STCN \cite{cheng2021rethinking} proposed to replace the memory encoder in STM with a lightweight encoder that does not require the mask as input. This improved both the speed and performance on the VOS task. 
For us, it implied that we would only need to run the encoder once. 
Thus, the speed of our system will be significantly faster without compromising performance if the system utilizes a STCN-type encoder, which we believe will be standard in the future.
In Fig. \ref{fig:run_time}, running VOS with STCN turns out to be at least $2.2\times$ faster for us than STM.

\section{Conclusion}
To generate better key frame proposals in unsupervised video multi-video object segmentation, we introduce a novel algorithm that aggregates object proposals in a local window to correlate space-time information in a graph. Maximal Cliques in this graph vote to eliminate false detections and ensure robust and accurate propagation of object proposals through a video sequence. With qualitative and quantitative experiments and analysis, we demonstrate that the mask proposal refinement provides significant performance improvements over state-of-the-art methods in the DAVIS-UVOS and Youtube-VIS benchmarks across different backbones and instance segmentation algorithms. In the future, we would like to pursue applications of this algorithm in realistic object discovery tasks, such as in robotics and autonomous driving applications.

{\small
\bibliographystyle{ieee_fullname}
\bibliography{main_arxiv}
}

\large{\textbf{Supplementary Materials:}}

\large{\textbf{A. Additional Run-time Analysis.}}
 The results ~in Table \ref{tab:davis-val-speed} are generated by using an NVIDIA Tesla V100 GPU. We report the detailed run-time of each module in \methodname.  It shows that our MCMPG framework is very lightweight and takes minimal time to run. In the same Table, we also show how the run-time changes as we increase the number of key frames. As mentioned in the main paper, with the STCN-style encoder, the tracker does not need to re-compute features from the backbone network for each track, making the proposed approach much faster compared to the case when the STM-style encoder was utilized. Also, in the case of STCN, the total run-time scales better with an increasing number of key frames. In Table \ref{tab:davis-val-speed}, we show the running time with STCN, which is at least $2.2\times$ faster even with $1$ key frame and more than $3.0 \times$ faster than with STM with more key frames.
 
 
 \begin{table}[ht!]
    \centering
    
    \caption{Run-time of {\methodname} on DAVIS-UVOS 2017 \textsl{val} with $1$, $2$, and $4$ key frames. While our framework is lightweight, when using STM  the speed suffers since STM requires re-running the backbone for each proposal. Alternatively, one could use newer tracking algorithms such as STCN where all the proposals can share the same backbone features, which would make the system much faster}
    \resizebox{1.0\linewidth}{!}{
    \begin{tabular}{l|r|r|r}
        \hline
        $Module$ & $1$ key frame & $2$ key frames  & $4$ key frames\\ \hline
        Candidate Generation & 7.50 & 14.60 & 29.52 \\
        Propagate to the key frame & 0.09 & 0.13 & 0.25\\
        {\graphnameshort} & 14.10 & 16.40 & 22.91\\
        Multi-Object tracking with STM & 475.60 & 971.10 & 1817.41\\

        Sequential NMS & 61.20 & 65.00 & 71.77\\ \hline
        
        Total time with STM  & 544.39 & 1,066.83 & 1941.86 \\ \hline \hline
         Multi-Object tracking with STCN & 182.51 & 261.93 &  378.63\\
   
         Total time with STCN & 251.29 & 357.66 & 503.02 \\
             &   \textbf{(2.2x faster)}  &  \textbf{(3.0x faster)} & \textbf{(3.9x faster)}\\ \hline
    \end{tabular}  }
    \label{tab:davis-val-speed}
\end{table}


\large{\textbf{B. Qualitative results}}

\begin{figure*}[ht!]
  \hspace{-0.2 in}
  \centering
  \includegraphics[width=\textwidth]{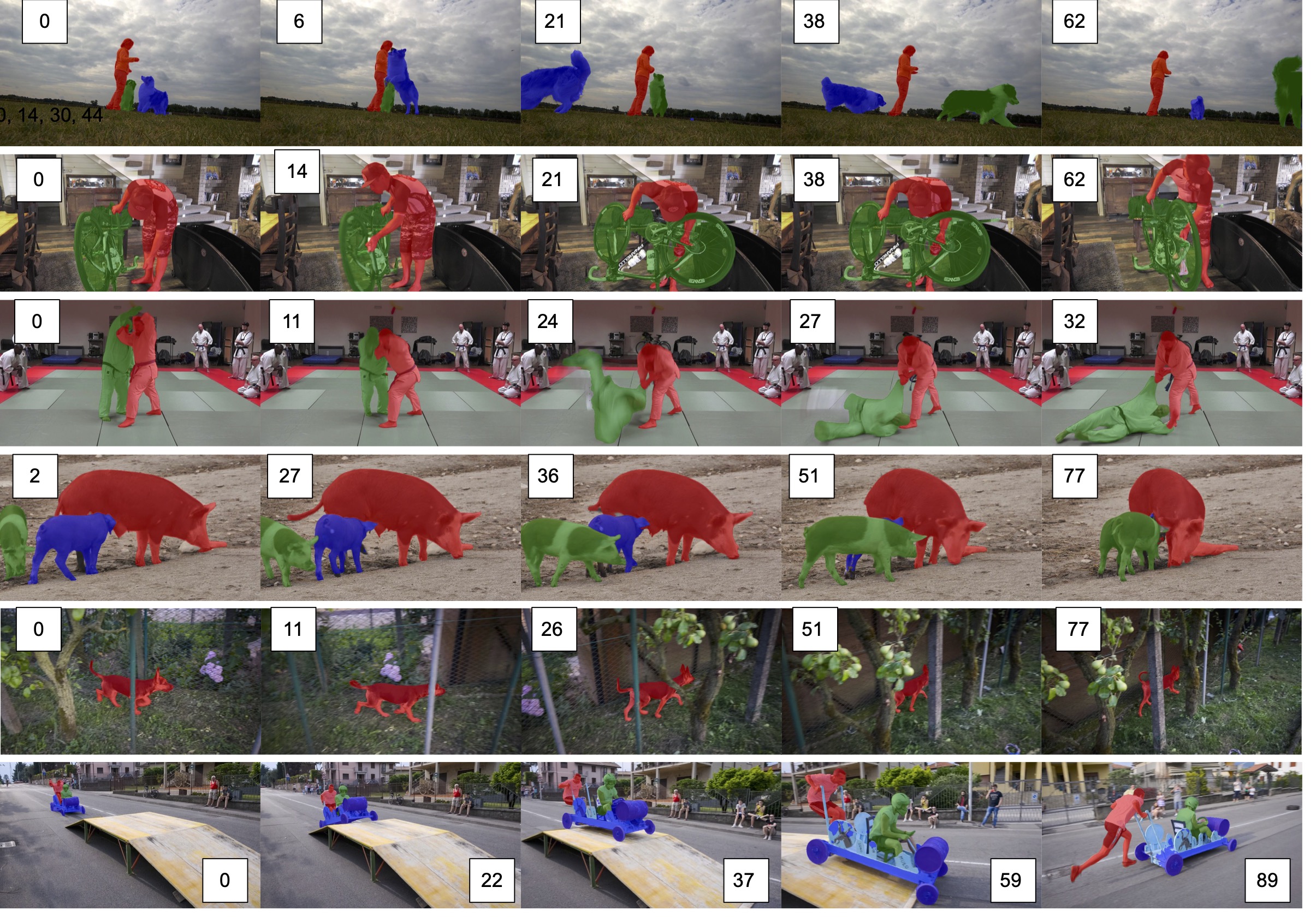}
  \caption { : Qualitative results on three sequences from the DAVIS 2017 \textsl{val} set. We show frames that are sampled from challenging scenarios such as fast motion, background clutter, occlusions, and multi-object interaction }
  \label{fig:qualitative}
  \vskip -0.25in 
\end{figure*}

\textbf{DAVIS 2017 \textsl{val}.} In Fig.~\ref{fig:qualitative}, we show the qualitative segmentation results of our approach on DAVIS-UVOS. The segmentation results are overlayed on the input RGB sequence where different colors are used to indicate different object instances. 

\begin{figure*}[ht!]
  \centering
  \hspace{-0.2 in}
  \includegraphics[width=\textwidth]{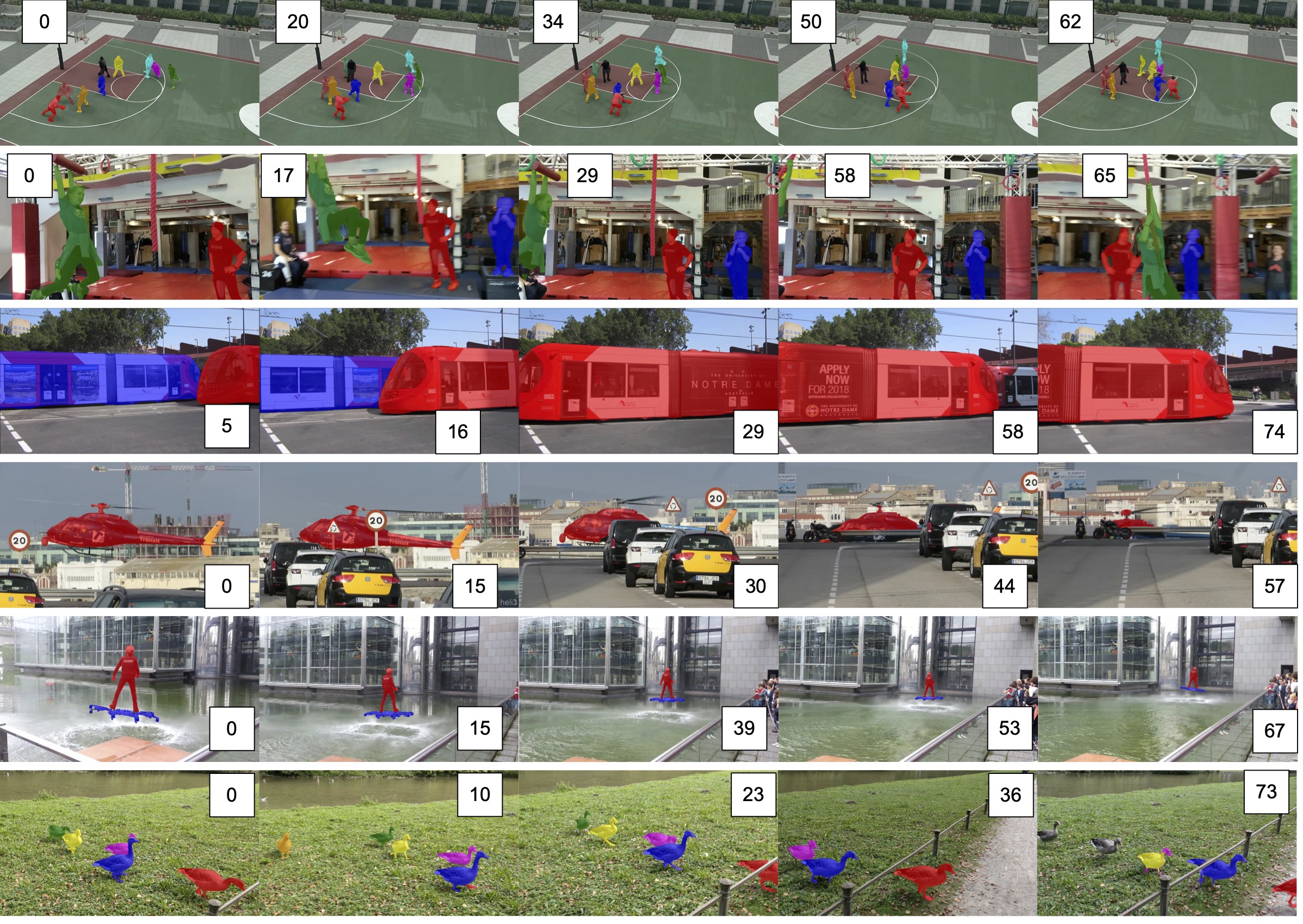}
  \vskip -0.1in
  \caption {Qualitative results on sequences from DAVIS 2019 \textit{test-dev}}
  \label{fig:test-davis-test-qualitative}
\end{figure*}

\textbf{DAVIS 2019 \textsl{test-dev}.} Some qualitative segmentation results of our approach on DAVIS-UVOS 2019 \textsl{test-dev} are shown in Fig.\ref{fig:test-davis-test-qualitative}.

\begin{figure*}[ht!]
  \centering
  \hspace{-0.2 in}
  \includegraphics[width=\textwidth]{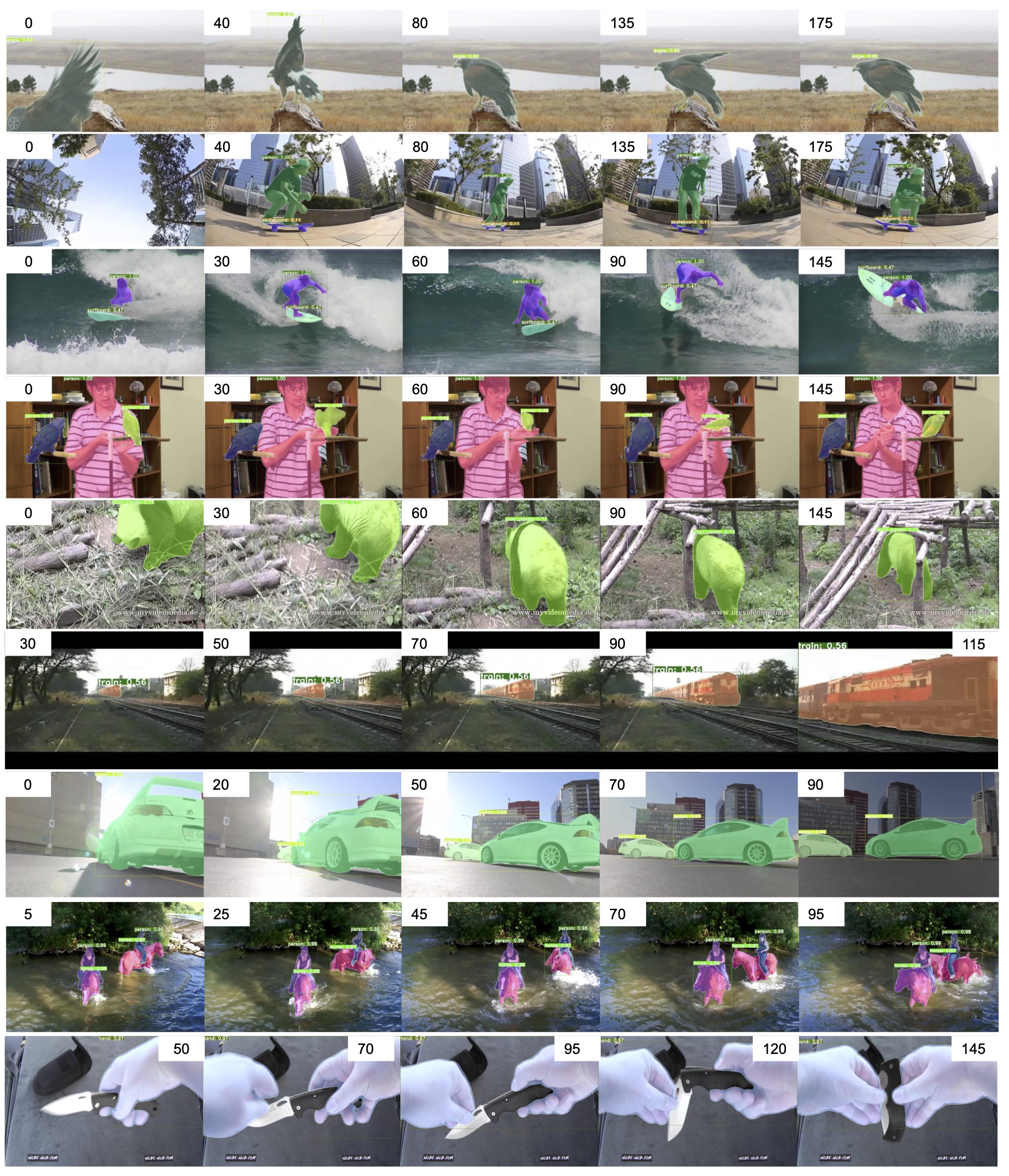}
  \caption {Qualitative results on sequences from YouTube-VIS 2019 \textit{val}}
  \label{fig:val-ytbvis-qualitative}
\end{figure*}

\textbf{YouTube-VIS 2019 \textsl{val}.} Some qualitative segmentation results of our approach on YouTube-VIS 2019 \textsl{val} are shown in Fig.\ref{fig:val-ytbvis-qualitative}.

\end{document}